\documentclass[runningheads]{llncs}

\usepackage{eccv}

\usepackage{eccvabbrv}

\usepackage{multicol}

\usepackage{multirow}

\usepackage{graphicx}
\usepackage{booktabs}
\usepackage{color}

\usepackage[accsupp]{axessibility}  

\usepackage[pagebackref,breaklinks,colorlinks,citecolor=eccvblue,linkcolor=eccvblue,urlcolor=eccvblue]{hyperref}

\usepackage{orcidlink}

\newcommand{\ql}[1]{\textcolor{black}{#1}}

\begin{document}

\title{Effective Adapter for Face Recognition in the Wild} 

\titlerunning{Abbreviated paper title}

\author{
Yunhao Liu$^{1}$,
~~~
Yu-Ju Tsai$^{2}$,~~~
Kelvin C.K. Chan$^{4}$,~~~
Xiangtai Li$^{3}$,~~~
Lu Qi$^{2}$\thanks{Corresponding author},~~~
Ming-Hsuan Yang$^{2,4}$, ~~~
\\[0.2cm]
$^1$Technical University of Munich~~
$^2$The University of California, Merced~~\\
$^3$Nanyang Technology University~~
$^4$Google Research~~
\\[0.2cm]  
\url{https://liuyunhaozz.github.io/faceadapter/}
}

\institute{}

\maketitle


\begin{abstract}
In this paper, we tackle the challenge of face recognition in the wild, where images often suffer from low quality and real-world distortions. 
Traditional heuristic approaches—either training models directly on these degraded images or their enhanced counterparts using face restoration techniques—have proven ineffective, primarily due to the degradation of facial features and the discrepancy in image domains. 
To overcome these issues, we propose an effective adapter for augmenting existing face recognition models trained on high-quality facial datasets. 
The key of our adapter is to process both the unrefined and enhanced images using two similar structures, one fixed and the other trainable. 
Such design can confer two benefits.
First, the dual-input system minimizes the domain gap while providing varied perspectives for the face recognition model, where the enhanced image can be regarded as a complex non-linear transformation of the original one by the restoration model. 
Second, both two similar structures can be initialized by the pre-trained models without dropping the past knowledge. 
The extensive experiments in zero-shot settings show the effectiveness of our method by surpassing baselines of about 3\%, 4\%, and 7\% in three datasets. 
Our code will be publicly available.
\end{abstract}    
\section{Introduction}
In recent years, deep learning
techniques have achieved remarkable progress in the field of face recognition~\cite{Cheng2019FaceSD, Wang2020HierarchicalPD, Kim2020GroupFaceLL, deng2019arcface, Wang2018CosFaceLM, Liu2017SphereFaceDH, Liu2019AdaptiveFaceAM}. 
%
%
However, face recognition in real-world scenarios~\cite{Hua2011IntroductionTT, Moeini2017OpensetFR, 9706804}, often termed as \textit{recognition in the wild}, still poses significant challenges \ql{due to the severe distortions and degradations of the used images}. 
%
%
%
\ql{Those images} significantly affect the effectiveness and reliability of face recognition systems, thereby hindering their broader deployment in practical applications.
%
%

\ql{An intuitive way to address this challenge is to directly train the face recognition models~\cite{Gu2022VQFRBF, Li2020EnhancedBF, Yang2021GANPE,tsai2024dual} using the collected low-quality images.}
\ql{However, the low-quality images often suffer from the problem of blurring or poor resolution and lack obvious clues even for human eyes.}
%
%
Another potential solution \ql{is employing a face restoration model to obtain the enhanced images.}
%
However, they would lose vital information from the original images, as restoration models do not consistently achieve perfect recovery. 
Therefore, retaining the original low-quality images in conjunction with the enhanced ones is essential, ensuring that the model remains the original image domain encountered in real-world settings \ql{but has more explicit features by the enhanced images}.

Based on the analysis above, we introduce an adapter framework designed to \ql{improve} face recognition in real-world (wild) conditions. 
The key to this framework is an adapter design integrated with a pre-trained face recognition model. 
Such a design harnesses the capabilities of existing face restoration models by applying the adapter to enhance high-quality images. 
Specifically, our framework employs a dual-branch structure: the adapter processes the restored images, and the frozen, pre-trained face recognition model handles the original low-quality images. 
After that, we propose a novel fusion module to ensemble the features of both two views by nested Cross- and Self-Attention mechanisms. 
Last, the fused features are used for similarity calculation. 

\begin{figure*}[tp!]
\centering
\includegraphics[width=0.9\linewidth]{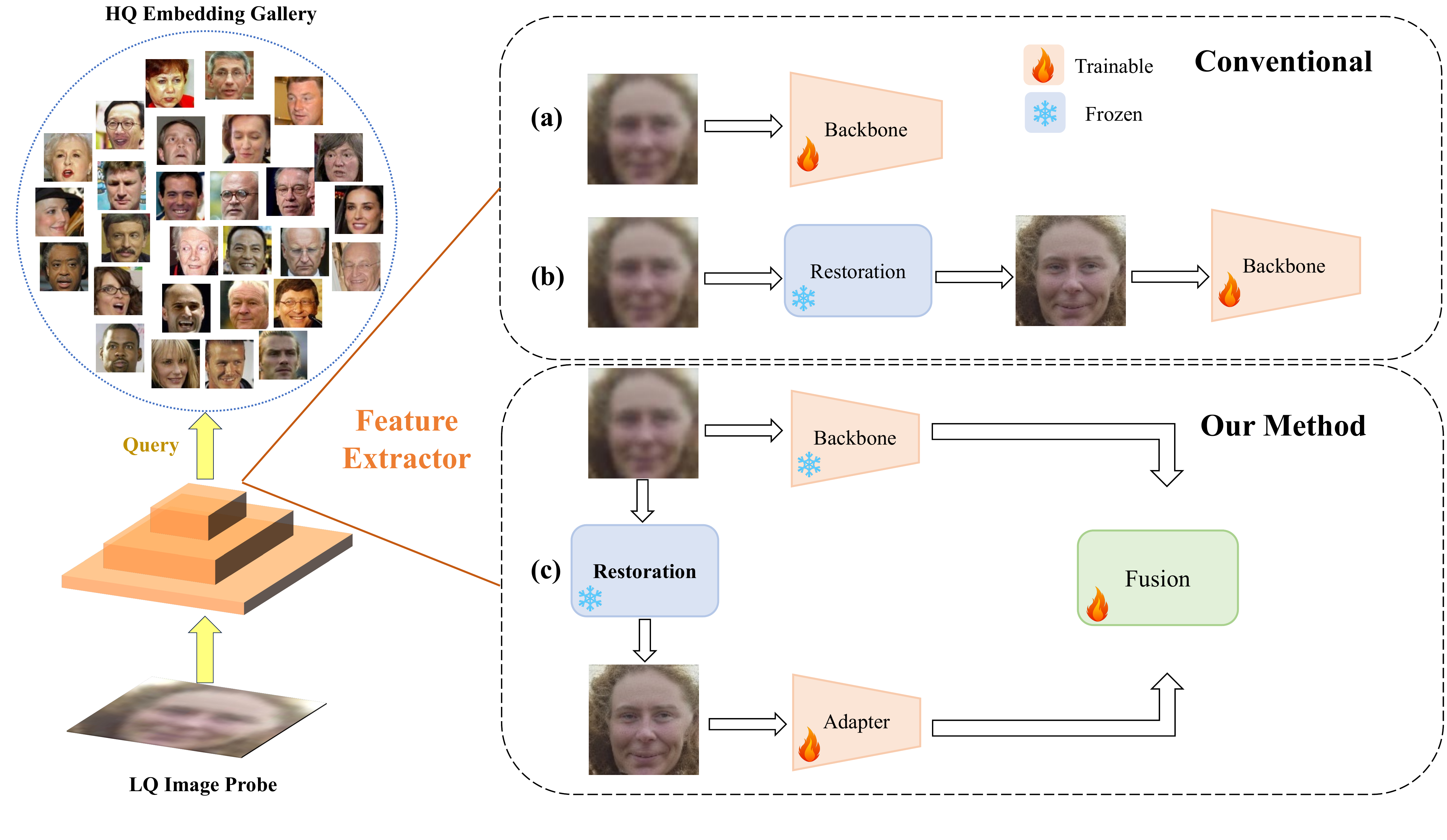}
\caption{In real-world applications, face recognition systems frequently encounter probe images of low quality (LQ), which presents a significant domain gap compared to the high-quality (HQ) embedding gallery. Our method (c) addresses this challenge by integrating the features from the LQ images with those of enhanced HQ images in the fusion structure. Compared with conventional methods (a) (b), our method effectively bridges the domain gap, ensuring more accurate and reliable face recognition performance in real-world conditions. \ql{In this figure, we can manifest the difference between past recognition and in-the-wild settings. So we have shown the difficulties and then propose our method. }} 
\label{fig:motivation}
\end{figure*}

We conduct extensive experiments \ql{in two settings, including} simulated and real-world environments commonly used for image restoration and domain adaptation.
\ql{For a simulated setting,} we train our model on one face recognition dataset and then test its effectiveness on the other three additional ones in a zero-shot manner. 
\ql{On the other hand, we train and test 7M large-scale low-quality images collected in the wild.}
\ql{Both settings} demonstrate the robustness and adaptability of our adapter framework in the wild. Fig.~\ref{fig:motivation} provides an overview of our approach.

%
Our main contribution is concluded as follows:
\begin{itemize}
\item We propose a novel framework for face recognition in the wild. 
It simultaneously processes low-quality and high-quality images enhanced by the face restoration model, bridging the gap between different image domains.
\item The key to our face recognition framework is an adaptor design that the pre-trained model on high-quality images can initialize. 
It allows the model to adapt low-quality images quickly without training from scratch.
In this way, the adaptor design keeps the original performance of low-quality images as the lower bound.
\item With the help of Cross-Attention and Self-Attention mechanisms, the extensive experiments show the considerable accuracy and reliability of the recognition process in the wild.
\end{itemize}

\section{Related Work}

\subsection{Face Recognition Methods}
Recent advancements in deep learning (DL) have significantly propelled the field of face recognition (FR). Historically, before the advent of deep learning, face recognition is reliant on manually designed features with methods such as Eigenface \cite{Belhumeur1996EigenfacesVF} and Fisherface \cite{anggo2018face}. The integration of deep learning has revolutionized this field, with a focus on model training and testing involving various network architectures. Recent algorithms in face recognition are categorized based on several factors, including the design of the loss function \cite{deng2019arcface, Wang2018CosFaceLM, Wang2018AdditiveMS, Zhang2019AdaCosAS, Liu2019FairLM, Sun2020CircleLA, Duan2019UniformFaceLD, Huang2020CurricularFaceAC}, as well as the representation and refinement of embedding \cite{ Hu2018PoseGuidedPF, Kim2020GroupFaceLL, Huang2021WhenAF, Cheng2019FaceSD, Wang2020HierarchicalPD}. Building on the advancements in deep learning for face recognition, our research specifically targets face recognition under low-quality image conditions.

\subsection{Face Restoration Methods}
Face Restoration (FR) aims to convert low-quality (LQ) facial images into high-quality (HQ) counterparts. Traditional methods relied on statistical priors and degradation models \cite{Baker2000HallucinatingF, Chang2004SuperresolutionTN, Li2006IlluminationTF, Wang2011FaceSS, Gao2012FaceSS} but often fell short in practical applications. The advent of deep learning has led to significant progress in this field. Recent research in deep learning-based \cite{Wang2021TowardsRB, Yang2021GANPE, Gu2022VQFRBF, wang2022restoreformer, zhou2022codeformer,tsai2024dual} face restoration methods includes a variety of techniques, network architectures, loss functions, and benchmark datasets. These advances are crucial in addressing challenges in face recognition. However, the Face recognition model trained with restored images suffers from domain gap problems in recognizing low-quality images. To address this, our model adopts a dual-input approach, integrating information from both low- and high-quality images. 


\subsection{Face Recognition in Low-Quality Images}
Face recognition in low-quality images has become increasingly important, especially in video surveillance and other real-world scenarios where capturing high-quality images may not be feasible. Research in this area has been categorized into four main approaches: (i) super-resolution based methods \cite{Wang2015DeepNF, Jiang2017NoiseRF, Farrugia2015FaceHU}, (ii) methods employing low-resolution robust features \cite{Ben2012GaitFC, Gao2018LowRankRA, Wang2016PoseRL, Herrmann2013ExtendingAL}, (iii) methods learning a unified representation space \cite{Biswas2012MultidimensionalSF, Ren2012CoupledKE, Li2010LowResolutionFR, Wang2015LowresolutionDF, AlMaadeed2016LowqualityFB}, and (iv) remedies for blurriness \cite{Shen2018DeepSF, Chrysos2017DeepFD, Flusser2016RecognitionOI, Jin2018LearningFD}. Most of these methods are non-deep learning based and typically focus either on low- or high-quality image data, but not both. In this paper, we propose an adapter for augmenting existing face recognition models.

\begin{figure*}[tp]
\centering
\includegraphics[width=\linewidth]{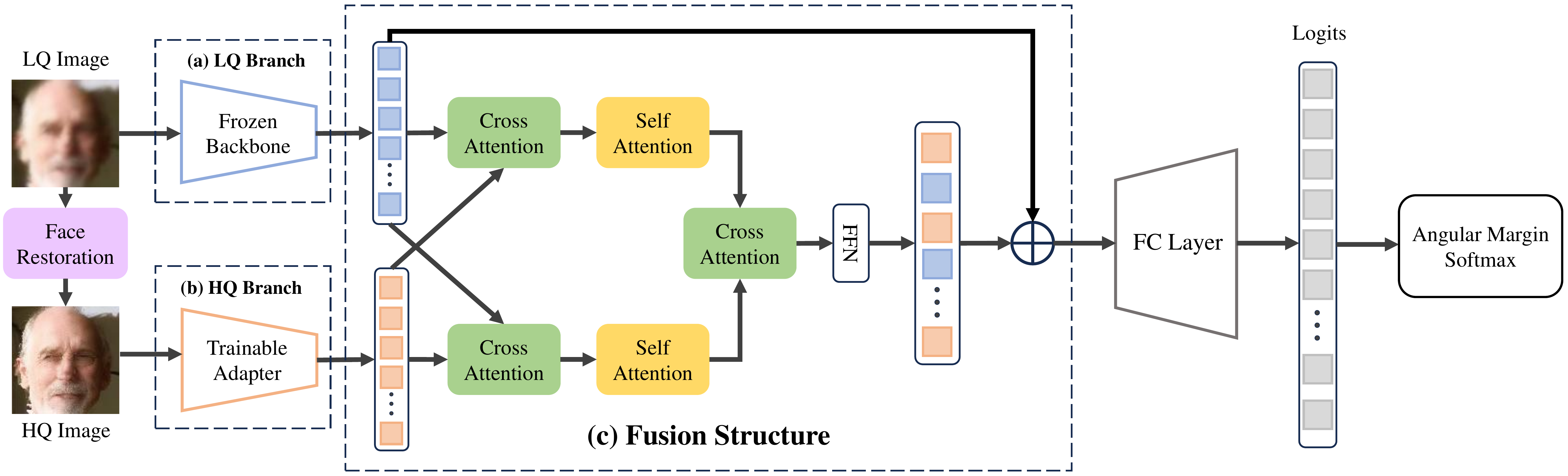}
\caption{\textbf{Joint Face Recognition Framework with Dual-Input Processing.} This architecture processes both low-quality (LQ) and restored high-quality (HQ) images, extracting them by two identical face recognition models. The Fusion Structure integrates feature sets before passing them to an Angular Margin Softmax function for loss computation, optimizing the network for enhanced recognition accuracy.} 
\vspace{-4mm}
\label{fig:network}
\end{figure*}
\section{Approach}
\ql{The motivation of adaptor design is to utilize images enhanced by the face restoration model. Compared to the original low-quality blurred images, the enhanced images can provide clearer facial features like eyes, eyebrows, etc. However, we are not able to directly drop the usage of original data due to the instability of the enhancement model.}
Therefore, we propose a \ql{face adaptor design}
that \ql{can effectively adapt the restored high-quality images to the existing face recognition models, as shown in Fig.~\ref{fig:network}.} 
%
\ql{At first, we freeze a well-trained face recognition model as our baseline. This baseline processes original low-quality images and maintains stable feature extraction.}
%
\ql{Our novelty design focuses on an adaptor that receives the enhanced images as another input source to complement the information missing in the original ones.}
\ql{Specifically, the adaptor has an HQ branch and fusion module where the HQ branch extracts the features of high-quality images, and the fusion module blends features extracted by image pairs.}
\ql{In the following subsections, we first briefly revisit the baseline face recognition model. Then, the adaptor design, including the HQ branch and fusion module, is introduced.}
%
%
%

\subsection{Baseline Model}



The baseline model aims at \ql{extracting robust} features from \ql{large-scale} low-quality images \ql{in the wild.} 
\ql{This setting is different from the traditional face recognition task that uses collected high-quality images for training and testing.}
\ql{Therefore, our baseline model should be trained by the low- 
 instead of high-quality images to decrease the domain gap.}
\ql{Considering there are few works in this area without available datasets to use, we explore two ways of generating 
low-quality images, including synthetic and real-world data, that will be introduced in Section~\ref{sec:data}.}



%
%
%

\ql{For the baseline model}, we choose Arcface~\cite{deng2019arcface} \ql{wrapped with} the ResNet-50~\cite{he2016deep} backbone without any specification. 
Given a low-quality image, we use the \ql{backbone} to extract the corresponding feature in, 
\begin{equation}
\mathcal{F}_f^l = \Phi_{l}(I^l), 
\end{equation}
where $\mathcal{F}_f\in \mathbb{R}^{B \times 512}$. $B$ is the image numbers in a batch, $I^l$ is the input low-quality images.



\subsection{Face Adaptor}
\ql{Our adaptor extracts the features of enhanced images to complement the original data. Thus, we design the HQ branch and fusion module to effectively extract robust features of enhanced images and fuse them into the original data.}

%
%

%

\subsubsection{HQ Branch}
This is an essential component of our framework, primarily focused on extracting features from high-quality images restored from their original, lower-quality versions. 
Similar to the \ql{backbone of the baseline model}, the HQ Branch \ql{mirrors an} ResNet-50 architecture \ql{that can be easily initialized by the pretrained weight}. 
%
%
%
%
\ql{In this way,} we provide a robust starting point for the \ql{adaptor's} training 
%
because that a well-established network's pre-learned patterns and features can lead to \ql{faster} convergence and improved performance. \ql{We represent the feature vector extracted by HQ branch as $\mathcal{F}_a \in \mathbb{R}^{B \times 512}$.}



%
%




\noindent \textbf{Fusion Structure}
%
\ql{We} employ a multi-layered structure of Cross-Attention and Self-Attention mechanisms to fuse the features of different sources. This nested fusion technique allows for a more nuanced and context-aware integration of features.



In this initial stage of the Fusion Structure, we focus on integrating features generated by the frozen backbone (\( \mathcal{F}_f \)) and the trainable adapter (\( \mathcal{F}_a \)).
We swap the positions of these two sets of features in the two identical Cross-Attention layers, respectively, followed by two identical Self-Attention Layers. 
This produces two sets of values, which are formulated as follows:

\begin{enumerate}
  \item [1)]  The first feature set, \( \mathbf{\mathcal{F}_{aff}} \), is defined by:

    \begin{equation}        
    \mathbf{\mathcal{F}_{aff}} = \mathbf{{SA_{\mathrm{1}}}} (\mathbf{{CA_{\mathrm{1}}}}(\underbrace{g_{\mathbf{q}}(\mathbf{\mathcal{F}_a})}_{\text{query}}, \underbrace{g_{\mathrm{k}}(\mathbf{\mathcal{F}_f})}_{\text{key}}, \underbrace{g_{\mathrm{v}}(\mathbf{\mathcal{F}_f})}_{\text{value}}))    \end{equation}

  \item [2)]  The second feature set, \( \mathbf{\mathcal{F}_{faa}} \), is given by:
    \begin{equation}
        \mathbf{\mathcal{F}_{faa}} = \mathbf{SA_{\mathrm{2}}}(\mathbf{CA_{\mathrm{2}}}(\underbrace{g_{\mathbf{q}}(\mathbf{\mathcal{F}_f})}_{\text{query}}, \underbrace{g_{\mathrm{k}}(\mathbf{\mathcal{F}_a})}_{\text{key}}, \underbrace{g_{\mathrm{v}}(\mathbf{\mathcal{F}_a})}_{\text{value}})) 
    \end{equation}
\end{enumerate}

Here, the statements $SA_{{\mathrm{1}, \mathrm{2}}}(\cdot)$ and $CA_{{\mathrm{1}, \mathrm{2}}}(\cdot)$ represent distinct instances of the same underlying Self-Attention and Cross-Attention model. 
$g_{\{\mathrm{q}, \mathrm{k}, \mathrm{v}\}}(\cdot)$ are the linear transformations to generating the query, key, and value in the attention process.
This stage is crucial for merging the distinct information from both low and high-quality images, enhancing the model's ability to recognize faces with greater accuracy and robustness.



%
%
\ql{After that}, we apply another single layer of Cross-Attention followed by a Feed Forward Network (FFN), further refining and fusing the features. This is mathematically represented as:

\begin{equation}
    \mathbf{\mathcal{F}_{fusion}} = \mathbf{FFN}(\mathbf{CA_{{\mathrm{3}}}}(\underbrace{g_{\mathbf{q}}(\mathbf{\mathcal{F}_{aff}})}_{\text{query}}, \underbrace{g_{\mathrm{k}}(\mathbf{\mathcal{F}_{faa}})}_{\text{key}}, \underbrace{g_{\mathrm{v}}(\mathbf{\mathcal{F}_{faa}})}_{\text{value}}))
\end{equation}

Here, $CA_{{\mathrm{3}}}$ is an instance of the Cross-Attention model. 
The utilization of \( \mathbf{\mathcal{F}_{aff}} \) as the query in this second stage is a strategic choice. 
It ensures that the fusion process continues to prioritize and effectively utilize the detailed information from the high-quality recovered images, which is essential for enhancing the accuracy and reliability of the final fused feature set for face recognition.

\ql{At last, we design a residual connection to }
enhance the robustness of the face recognition system by incorporating stable features from the Face Freeze Model with the dynamic features from the Fusion Structure. 
The mathematical formulation of the Residual Structure is as follows:

\begin{equation}
    \mathcal{R} = \mathcal{F}_f + \mathcal{F}_{fusion}
\end{equation}

Where \( \mathcal{R} \) represents the resultant residual feature, \( \mathcal{F}_f \) is the feature vector from the LQ Branch, and \( \mathcal{F}_{fusion} \) symbolizes the fused feature vector generated by the nested attention mechanisms mentioned before. 
The LQ Branch, characterized by its frozen, pre-trained backbone, provides a stable and consistent feature set. 
Adding these features to the fusion output ensures that the final feature representation retains a baseline level of quality and reliability.

%

\subsection{Training Objective}
\noindent\textbf{Angular Margin Softmax.} 
In this section, we elaborate on the Angular Margin Softmax \cite{deng2019arcface} loss function, which is integral to our model's ability to distinguish between different faces effectively. 

\begin{equation}
    \resizebox{0.9\hsize}{!}{
    $
    L = -\frac{1}{N} \sum_{i=1}^{N} \log \frac{e^{s \cdot \cos(m_1 \cdot \theta_{y_i} + m_2) - m_3}}{e^{s \cdot \cos(m_1 \cdot \theta_{y_i} + m_2) - m_3} + \sum_{j=1, j \neq y_i}^{n} e^{s \cdot \cos \theta_j}}
    $
    }
\end{equation}
%
In this formula, \( \theta_{y_i} \) is the angle between the feature vector and the class center for the correct class \( y_i \), and \( N \) represents the number of classes. \ql{In our experiments, the default setting} of parameters is \( m_1 = 1.0 \), \( m_2 = 0.5 \), \( m_3 = 0.0 \), and \( s = 64 \). 
%


 \section{Experiments}

\subsection{Datasets}
\label{sec:data}
\noindent\textbf{Synthetic Datasets.} 

Our experiments utilize the UMDFaces~\cite{bansal2017umdfaces} as our training dataset, with $367,888$ images across $8,277$ labels, and we test on three public datasets: LFW~\cite{huang2008labeled} with $12,000$ images, CFP-FP~\cite{sengupta2016frontal} with $14,000$ images, and AgeDB~\cite{moschoglou2017agedb} with $12,000$ images. To simulate the degradation of the real-world environment, we incorporate a simulation tool, TurbulenceSim~\cite{chimitt_chan_sim}, to apply different levels of degradation to our training and testing datasets. We also want to note that atmospheric turbulence is a more challenging and complex degradation type.



We use different levels of degradation called $10k$, $20k$, $30k$, and $40k$, representing the turbulence intensity in the TurbulenceSim~\cite{chimitt_chan_sim}. Larger values indicate greater image blurring and turbulence effects. Using this algorithm, we can apply varied degradation effects to the face images in our dataset, allowing for a controlled experimentation environment. 

\noindent\textbf{Real-world Datasets.}

To validate the effectiveness of our method, we use the BRIAR~\cite{cornett2023expanding} dataset as our real-world training and testing dataset. The BRIAR dataset contains gallery images under controlled environments and probe images under challenge scenarios, which captured distances ranging from 100m to 1000m. The main degradation type of this dataset is atmospheric turbulence, which is a more complex and challenging degradation. We use a subset containing 7M images and divide the subset with 5M images for training and 2M images for testing.

\subsection{Implementation Details}
\noindent\textbf{Evaluation Metrics.}

We perform zero-shot face verification using our model across three datasets: LFW~\cite{huang2008labeled}, CFP-FP~\cite{sengupta2016frontal}, and AgeDB~\cite{moschoglou2017agedb}. We adopt a $1:1$ face recognition verification accuracy, where the verification set images are divided equally into gallery and probe sets. We use the pre-trained ArcFace~\cite{deng2019arcface} backbone to generate embeddings for gallery images. For the probe images, we use our framework to generate the embeddings.

For the BRIAR~\cite{cornett2023expanding} dataset, we follow the evaluation protocol provided by the dataset, and the main evaluation metric is TAR@0.01FAR.



\noindent\textbf{Experimental Settings.}

Our Joint Framework comprises two parallel branches: the LQ and HQ branches. Both branches take the ArcFace~\cite{deng2019arcface} backbone as the pre-trained weight. We feed the LQ facial image to our LQ Branch, and we use the CoderFormer~\cite{zhou2022codeformer} model with the fidelity weights $w=0.5$ to generate HQ facial images for the HQ Branch.


The training is conducted with a batch size of $256$ across $5$ epochs, processing images at a resolution of $112\times112$. We optimize the model using SGD with an initial learning rate of $0.02$, momentum set to $0.9$, and weight decay of $5e^{-4}$. The learning rate follows a Polynomial Warmup schedule, which starts with a warmup phase where the learning rate ramps up, followed by a polynomial decay that adjusts the learning rate based on the progress of training epochs. The LQ Branch remains frozen throughout this process, while the HQ Branch and the Fusion Structure, including Self-Attention, Cross-Attention, and the Feed Forward Network, are actively trained.

\begin{table}[tp]
\centering
\small
\caption{Verification results (\%) of different face recognition models on LFW~\cite{huang2008labeled}, CFP-FP~\cite{sengupta2016frontal}, and AgeDB~\cite{moschoglou2017agedb} datasets under 20k degradation intensity. For methods ArcFace~\cite{deng2019arcface}, CosFace~\cite{Wang2018CosFaceLM}, and QAFace~\cite{Saadabadi_2023_WACV}, we directly test LQ images on their pre-trained model. For our method, we use both LQ and HQ images as input.}
\begin{tabular}{cccc}
\hline Methods  & LFW (20k) & CFP-FP (20k) & AgeDB (20k)\\
\hline ArcFace~\cite{deng2019arcface} (0.5) & 98.500 & 90.157 & 92.350  \\
\text CosFace~\cite{Wang2018CosFaceLM} (0.35) & 99.033 & 93.371 &93.117  \\
\text QAFace~\cite{Saadabadi_2023_WACV} (0.5) & 98.317 & 90.629 & 80.667 \\
\hline \textbf{Ours} (ArcFace) & \textbf{99.133} &\textbf{91.529} & \textbf{93.850} \\
\text {\textbf{Ours} (CosFace)} &\textbf{ 99.400} & \textbf{93.886} & \textbf{94.317}\\
\text {\textbf{Ours} (QAFace)} & \textbf{99.167} & \textbf{93.314} & \textbf{87.350}\\
\hline
\end{tabular}
\vspace{-4mm}
\label{tab:main_111}
\end{table}

\subsection{Comparison with State-of-the-art}

\noindent\textbf{Synthetic Datasets.} 

As shown in Table \ref{tab:main_111}, we demonstrate the effectiveness of our model in enhancing the performance of standard face recognition models like ArcFace~\cite{deng2019arcface}, CosFace~\cite{Wang2018CosFaceLM}, and QAFace~\cite{Saadabadi_2023_WACV}, especially under conditions of 20k degradation intensity on the LFW~\cite{huang2008labeled}, CFP-FP~\cite{sengupta2016frontal}, and AgeDB~\cite{moschoglou2017agedb} datasets. The experimental results show that when our approach is applied to these models, there is an improvement in their ability to recognize faces in LQ images. This improvement is attributed to our model's capability to effectively process and combine features from LQ and HQ images, thereby enriching the existing face recognition models with an increased capacity to handle image quality degradation. This experiment highlights our approach's potential ability as a versatile tool for improving face recognition accuracy in real-world scenarios with LQ images.

\noindent\textbf{Real-world Datasets.} 

To evaluate the practical applicability of our proposed framework, we extend our analysis to the real-world dataset BRIAR~\cite{cornett2023expanding}, which is introduced earlier. This dataset presents a challenging environment for face recognition due to its inclusion of atmospheric turbulence as the main degradation type. Given the complex and unpredictable nature of real-world conditions, the BRIAR dataset serves as an ideal benchmark for assessing the efficiency of our model. Employing the same methodology applied to the Synthetic Datasets, we process the Real-world Datasets, focusing on overcoming the challenges posed by atmospheric turbulence. Our framework demonstrates a performance improvement.

Table~\ref{tab:real_world_performance} shows the performance metrics before and after applying our model to the real-world dataset.

\begin{table}[tp]
\centering
\caption{Face recognition performance (TAR@0.01FAR) on the BRIAR~\cite{cornett2023expanding} Dataset. For the QAface method We directly test LQ images on its pre-trained model. Our method uses the QAface model as our Baseline Model and uses both LQ and HQ images as input.}
\label{tab:real_world_performance}
\begin{tabular}{cc}
\hline
\textbf{Methods} & \textbf{TAR@0.01FAR}  \\
\hline
QAFace~\cite{Saadabadi_2023_WACV} & 0.637  \\
\textbf{Ours} (QAFace) & \textbf{0.638} \\
\hline

\end{tabular}
\end{table}









\subsection{Ablation Study}


In this section, we conduct comprehensive ablation studies to evaluate the effectiveness of different components in our face recognition system. The ablation studies contain the effect of image degradation, the performance of backbone models under quality variations, the efficacy of feature fusion methods, and the strategic implementation of attention mechanisms. This collective analysis is crucial for understanding and enhancing the system's resilience and accuracy in diverse conditions.




\begin{figure*}[tp]
  \begin{subfigure}{\linewidth}
  \includegraphics[width=.33\linewidth]{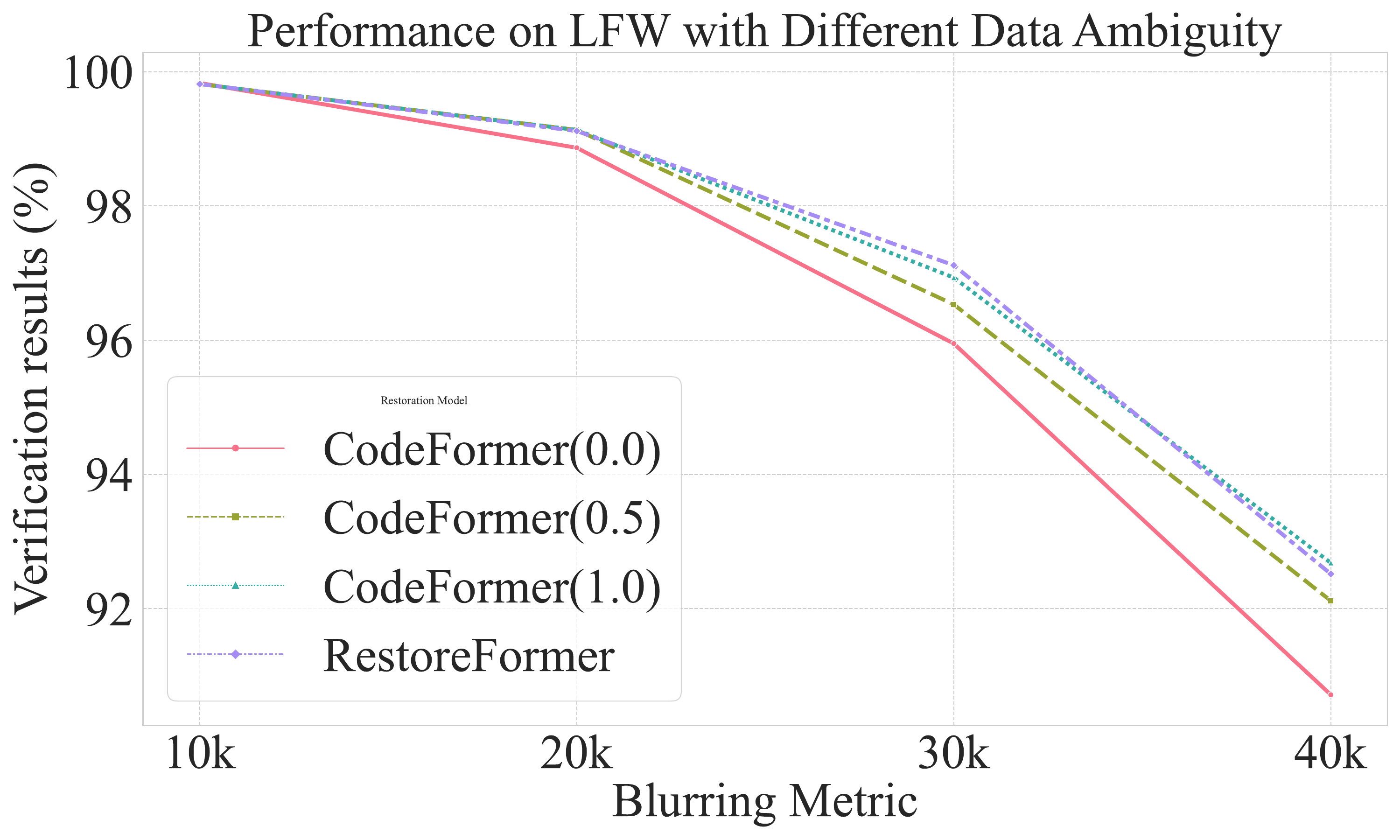}\hfill
  \includegraphics[width=.33\linewidth]{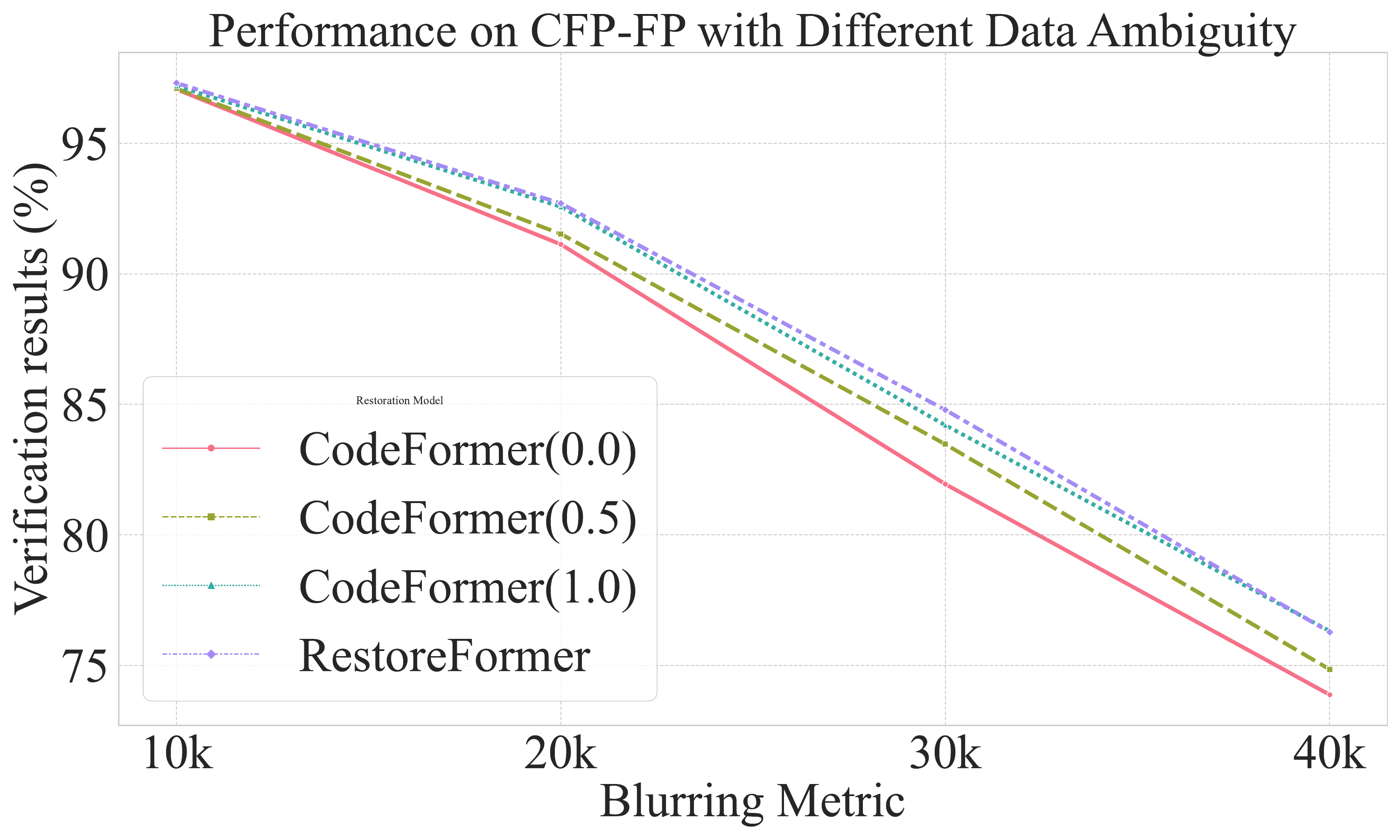}\hfill
  \includegraphics[width=.33\linewidth]{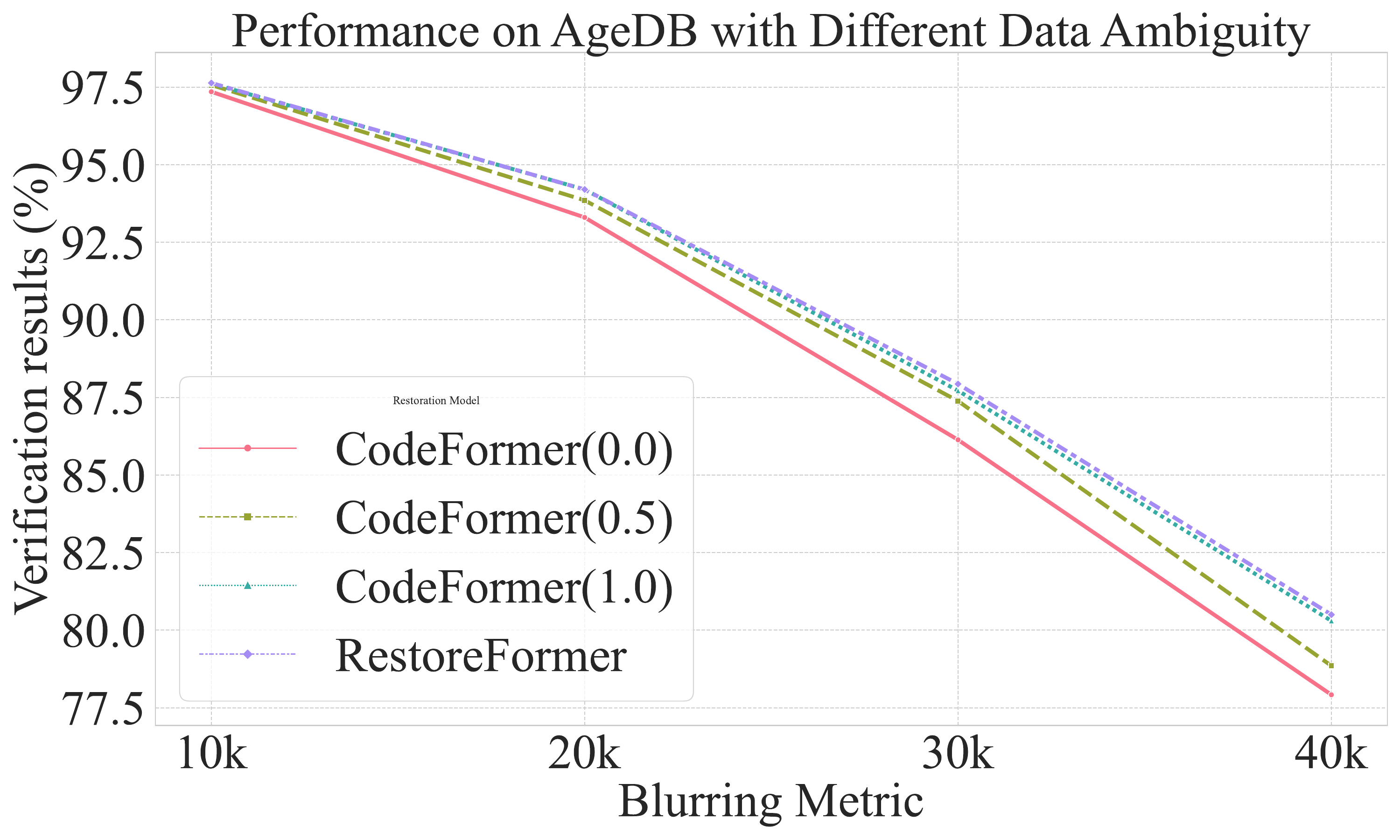}
  \end{subfigure}

  \caption{Verification performance (\%) using different face restoration methods on LFW, CFP-FP, and AgeDB with different Degradation Intensity. 
  }
\vspace{-4mm}
  \label{fig: Face Restoration}
\end{figure*}

\begin{table}[tp]
\centering
\small
\caption{Verification results (\%) of the model using different face recognition pre-trained backbone on LFW~\cite{huang2008labeled}, CFP-FP~\cite{sengupta2016frontal}, and AgeDB~\cite{moschoglou2017agedb} datasets with 20k degradation intensity.}
\begin{tabular}[width=\linewidth]{cccc}
\hline Backbones  & LFW (20k) & CFP-FP (20k) & AgeDB (20k)\\
\hline ArcFace~\cite{deng2019arcface} & 99.133 &91.529 & 93.850 \\
\text CosFace~\cite{Wang2018CosFaceLM} & 99.400 & 93.886 & 94.317\\
\text QAFace~\cite{Saadabadi_2023_WACV} & 99.167 & 93.314 & 87.350\\
\hline
\end{tabular}
\label{tab: face_rec}
\end{table}


\noindent \textbf{Compare face restoration methods with different degradation levels.} 
In our designed system, we first need to use face restoration methods to restore the LQ images to HQ images. We conduct ablation studies to verify the impact of different face restoration models with different levels of degraded data.
We choose CodeFormer~\cite{zhou2022codeformer} and RestoreFormer~\cite{wang2022restoreformer} as our restoration models for comparison. Specifically, within the CodeFormer framework, we configure three variants with varying image restoration effects by adjusting the fidelity weights $w=0.0,0.5,1.0$. These variations are utilized in the subsequent ablation Studies to evaluate their performance differences.

As shown in Fig.~\ref{fig: Face Restoration}, we evaluate these face restoration models on testing datasets with $10k$, $20k$, $30k$, and $40k$ degradation intensities. The experimental results show that using RestoreFormer as the restoration model in our method has the best face recognition accuracy. Notably, CodeFormer's recognition accuracy improves as the fidelity weight $w$ increases. Consistent with expectations, a significant decrease in face recognition accuracy correlates with increasing degradation levels.

\begin{table}[tp]
\centering
\small
\caption{Verification results (\%) using four different face recognition strategies on LFW, CFP-FP, and AgeDB Datasets with 20k degradation intensity. We use the same ArcFace pre-trained backbone.
  }
\begin{tabular}{lccc}
\hline Methods  & LFW (20k) & CFP-FP (20k) & AgeDB (20k)\\
\hline  w/o Restoration  & 98.500 & 90.157 &  92.350 \\
  After Restoration  & 91.517 & 79.029 &  80.750 \\
Fine-tuning & 98.717 & 89.014 & 86.700 \\
\hline
\textbf{Ours} & \textbf{99.133} & \textbf{91.529} & \textbf{93.850}\\
\hline
\end{tabular}
\vspace{-4mm}
\label{tab: Comparative Analysis}
\end{table}

\noindent \textbf{Different pre-trained face recognition backbone.} 
The backbone model in our face recognition system serves as the foundational architecture upon which the performance of the entire system is heavily reliant. In our experiments, we choose to evaluate three prominent backbone models: ArcFace~\cite{deng2019arcface}, CosFace~\cite{Wang2018CosFaceLM}, and QAFace~\cite{Saadabadi_2023_WACV}. These models are selected based on their distinct approaches to handling face recognition tasks and their reported efficacy in existing literature.

As shown in Table~\ref{tab: face_rec}, our model with the CosFace backbone has achieved the highest accuracy among all tested methods in the testing datasets, with its performance being especially prominent in the AgeDB dataset. This indicates that this particular version of the algorithm is highly effective in face recognition tasks involving LQ facial images.

\noindent \textbf{Different strategies to train and evaluate face recognition model.}
In this ablation study, we conduct experiments about the training strategies under various input image conditions.
As shown in Table~\ref{tab: Comparative Analysis}, we compare four training and evaluation strategies for face recognition models. For these four experiments, we all use ArcFace~\cite{deng2019arcface} as our pre-trained backbone.

\begin{itemize}
  \item \textbf{Evaluation with original images:} Given original images, we use the pre-trained face recognition model to compute the face embedding without any training.

  \item \textbf{Evaluation with restored images:} 
  Given restored images, we use the pre-trained face recognition model to compute the face embedding without any training.
    
  \item \textbf{Fine-tune with restored images:} We first fine-tune the pre-trained recognition model with the restored images, and we then pass the restored testing image to the model to generate the embedding. 
  \item \textbf{Our Method:} We use our structure to jointly train with original and restored images. Given original and restored images, we fuse the features from two branches and generate the final embedding.
\end{itemize}

As shown in Table~\ref{tab: Comparative Analysis}, the experimental results show that performance drops significantly when directly using the restored images as the input for the pre-trained model, which indicates there is a significant domain gap between the original and restored images.
Fine-tuning the model on restored images does improve as the model learns the properties of the restored images. However, the domain gap problem still exists because the model lacks information about the original images after training.
Our fusion structure approach significantly outperforms other methods, which combine features extracted from original and restored images.


\noindent \textbf{Residual structure inside fusion structure.} We conduct the ablation study to examine the effectiveness of the residual structure inside our fusion structure. The residual structure is designed to integrate the embedding from the LQ Branch with the fused feature obtained from our Fusion Structure.

As shown in Table~\ref{tab: Residual Structure in Fusion}, the experimental results show higher face recognition accuracy with residual structure in the LFW and AgeDB datasets, and the increase is more pronounced. The results indicate that the residual structure plays a role in our model. In addition, the residual structure utilizes the features generated by the pre-trained, frozen backbone. This integration ensures the baseline quality of the feature.

\begin{table}[tp]
  \centering
  \small
  \caption{Verification results (\%) of models with and without residual structure in LFW, CFP-FP, and AgeDB datasets at 20k degradation intensity.}
\begin{tabular}{lccc}
\hline Methods  & LFW (20k) & CFP-FP (20k) & AgeDB (20k)\\
\hline w/o Residual & 98.567 & \textbf{91.614} & 93.383\\
With Residual \textbf{(Ours)} & \textbf{99.133} & 91.529 & \textbf{93.850}\\
\hline
\end{tabular}
\vspace{-4mm}
\label{tab: Residual Structure in Fusion}
\end{table}

\begin{table}[tp]
  \centering
\small
\caption{Verification performance (\%) using different cascade layers of fusion structure on LFW, CFP-FP, and AgeDB with 20k degradation intensity.}
\begin{tabular}{lccc}

\hline Methods & LFW (20k)&CFP-FP (20k)  & AgeDB (20k)\\
\hline Cascade $\times 5$ & 50.800 & 50.071 & 50.000 \\
\text Cascade $\times 3$ & 99.100& 91.329 & 93.317\\
\text No Cascade \textbf{(Ours)} & \textbf{99.133} &\textbf{91.529}  & \textbf{93.850}\\
\hline
\end{tabular}
\label{tab: Cascade Structure in Fusion}
\end{table}

\noindent \textbf{Cascade structure.}  We also explore the design of the cascade structure, which involves repeating our feature fusion method in successive stages. We hypothesize that a multi-layered fusion might enhance recognition accuracy by iteratively refining features.

Contrary to our hypothesis, as shown in Table~\ref{tab: Cascade Structure in Fusion}, our experiments show that cascading the fusion process five and three times does not improve results but reduces performance compared to our primary fusion method. This suggests that increasing the complexity with additional fusion stages does not necessarily benefit the face recognition system and that a single-stage fusion is sufficient for optimal performance. These outcomes highlight the delicate balance between model complexity and efficacy, advocating for simplicity when additional layers do not contribute to performance gains.


\begin{figure}[tp]
\centering
\includegraphics[width=\linewidth]{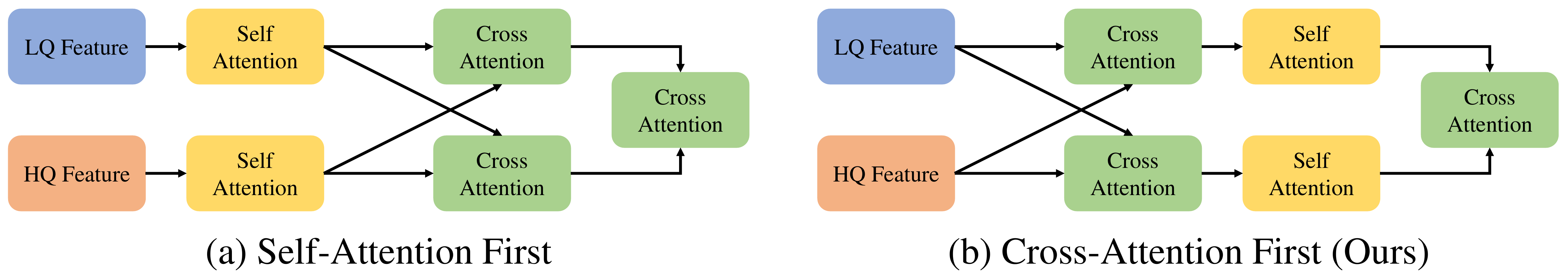}
\caption{Comparison between the two different attention orders in Fusion Structure. Our method uses (b) Cross-Attention First.}
\vspace{-4mm}
\label{fig:attention_order}
\end{figure}

\begin{table}[tp]
\centering
\small
\caption{ 
Comparison of face recognition accuracy (\%) between Cross-Attention First (\textbf{Ours}) method and Self-Attention First method across LFW, CFP-FP, and AgeDB datasets at 20k degradation intensity. }
\begin{tabular}{cccc}
\hline Methods  & LFW (20k) & CFP-FP (20k) & AgeDB (20k)\\
\hline 
Self-Attention First & 99.100 & 91.500 & 93.583\\
Cross-Attention First \textbf{(Ours)} & \textbf{99.133} & \textbf{91.529} & \textbf{93.850}  \\
\hline
\end{tabular}
\label{tab: Orders of Attention Mechanisms.}
\end{table}

\noindent \textbf{Orders of Cross and Self-Attention mechanisms.} 
Fig.~\ref{fig:attention_order} shows two different orders of Self-Attention and Cross-Attention within our fusion structure, which processes parallel lines of LQ and HQ features.

\begin{itemize}
  \item \textbf{Self-Attention First:} We first pass the features with the Self-Attention layer independently. This is succeeded by the Cross-Attention layer to fuse the features from two branches. Next, we apply the additional Cross-Attention layer to integrate the final fusion feature.
  \item \textbf{Cross-Attention First (Ours):} We first pass the features with the Cross-Attention layer to generate the fused features. We then pass the fused feature to the Self-Attention layer to enhance the representation of the features. Next, we apply the additional Cross-Attention layer to integrate the final fusion feature.
\end{itemize}

As shown in Table~\ref{tab: Orders of Attention Mechanisms.}, the experimental results indicate that both methods achieve remarkably similar performance across all testing datasets. The close proximity in the results of the two methodologies suggests that our model's nested attention mechanism efficiently captures and utilizes the necessary information for face recognition, regardless of the order of the Cross-Attention and Self-Attention layers. 

This observation implies that the ability of our model to process and integrate features effectively is not significantly influenced by the alteration in the sequence of attention mechanisms. Consequently, this finding underscores the robustness and flexibility of our nested attention approach, demonstrating its capability to maintain high performance even when the order of attention layers is varied.

\begin{figure}[tp]
\centering
\includegraphics[width=\linewidth]{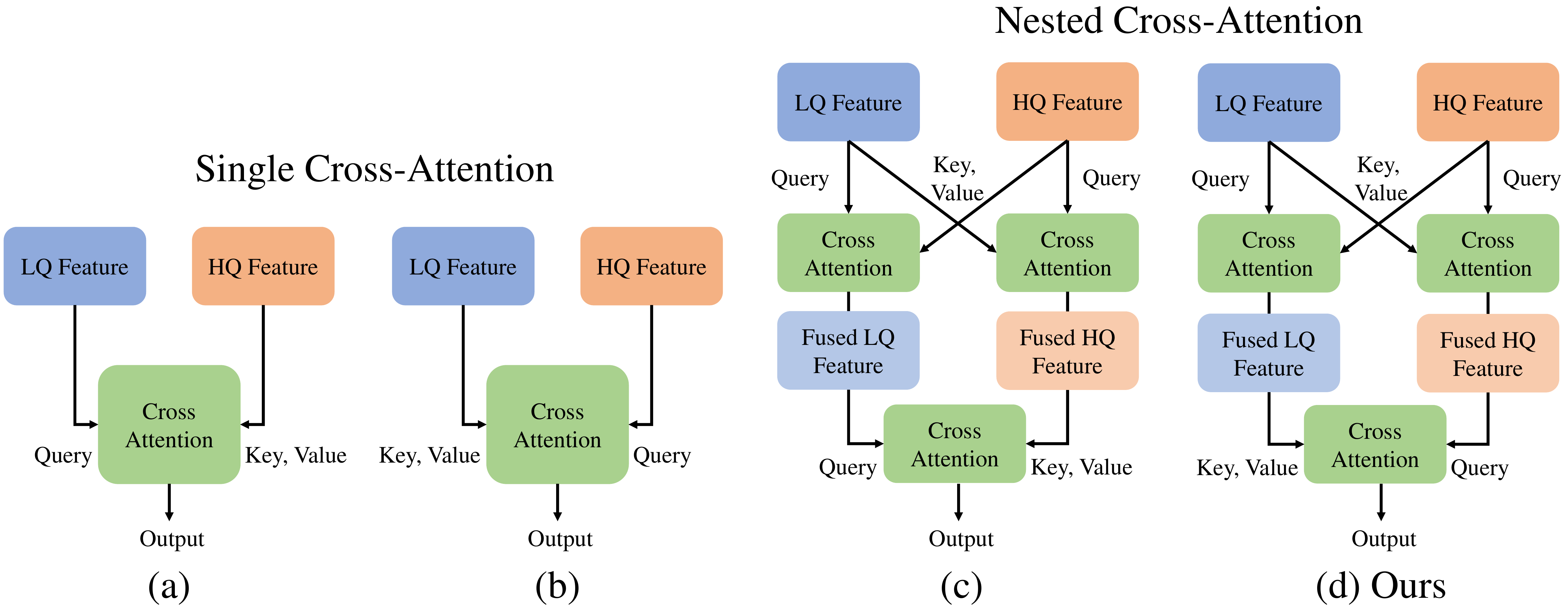}
\caption{Comparison of methods for different orders of input in Cross-Attention: (a), (b) are the methods for Feature as Query or Key and Value in single Cross-Attention input. (c), (d) are methods in nested Cross-Attention.
} 
\label{fig:keyvalue}
\end{figure}

\begin{table}[tp]
  \centering
  \small
  \caption{Face recognition accuracy comparison of different parameters in Cross-Attention across LFW, CFP-FP, and AgeDB Datasets at 20k degradation intensity. }
\begin{tabular}{cccc}
\hline Methods  & LFW (20k)  & CFP-FP (20k) & AgeDB (20k) \\
\hline (a) & 98.883 & 90.286 & 92.650 \\
 (b) & 99.017 & 91.557 & 93.583 \\
 (c) & 98.683 & 89.929 & 92.300  \\
 (d) \textbf{(Ours)} & \textbf{99.133} & \textbf{91.529} &  \textbf{93.850} \\
\hline
\end{tabular}
\label{tab: Parameters in Cross Attention}
\end{table}

\begin{figure}[tp]
  \begin{subfigure}{\linewidth}
  \includegraphics[width=.22\linewidth]{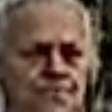}\hfill
  \includegraphics[width=.22\linewidth]{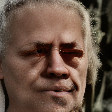}\hfill
  \includegraphics[width=.22\linewidth]{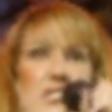}\hfill
  \includegraphics[width=.22\linewidth]{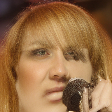}
  \caption{Two sets of low-quality images and the restored high-quality images using CoderFormer (w=0.5).}
  \end{subfigure}\par\medskip
  \begin{subfigure}{\linewidth}
  \includegraphics[width=.22\linewidth]{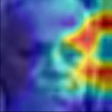}\hfill
  \includegraphics[width=.22\linewidth]{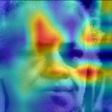}\hfill
  \includegraphics[width=.22\linewidth]{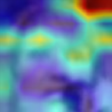}\hfill
  \includegraphics[width=.22\linewidth]{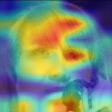}
  \caption{Heat map corresponding to the image in (a), aligned by columns.}
  \end{subfigure}\par\medskip
  \caption{We use GramCAM~\cite{jacobgilpytorchcam} for feature visualization of low-quality images and recovered images fed into the face recognition model.}

  \label{fig:heat map}
\end{figure}

\noindent \textbf{Different number and order of Cross-Attention.} We conduct ablation studies on the different orders of Cross-Attention to discern the optimal configuration of the Query, Key, and Value roles within the Cross-Attention layer that processes high and low-quality image features.

In the Cross-Attention modules of our Unified Feature Integration Module, we design different variations of structures, as shown in Fig.~\ref{fig:keyvalue}, containing designs of single Cross-Attention and nested Cross-Attention. We design four experimental settings to fuse HQ and LQ features.

\begin{itemize}
  \item(a) For single Cross-Attention design, we use LQ features as Query and HQ features as Key and Value.
  \item(b) Similar to setting (a), we use HQ features as Query and LQ features as Key and Value.
  \item(c) For nested Cross-Attention design, we first apply two Cross-Attention layers to obtain the fused HQ and LQ features. We then use the Fused LQ features as Query, and the Fused HQ features as Key and Value.
  \item(d) \textbf{(Ours)} Similar to setting (c), we use the Fused HQ features as Query and the Fused LQ features as Key and Value.
\end{itemize}

Table \ref{tab: Parameters in Cross Attention} shows the results of these experiments on the testing datasets.
The experimental results indicate that design (d) has the best recognition scores on the test datasets. This outcome suggests that using the Fused HQ features as the base for the Query in the final Cross-Attention layer is more effective, possibly due to the richer and more discriminative information inherent in the HQ features being leveraged to guide the fusion process. As shown in Fig. \ref{fig:heat map}, the face information of the restored HQ image is easier to extract by the face recognition model.

\section{Conclusion}
In this paper, we introduce a novel adapter framework to enhance face recognition in real-world scenarios with low-quality images. Our approach leverages a frozen pre-trained model and a trainable adapter to bridge the gap between original and enhanced images. Specifically, the Fusion Structure integrates advanced nested Cross-Attention and Self-Attention mechanisms. The extensive experiments across multiple datasets show that our method significantly improves accuracy and reliability in face recognition than conventional techniques. This work sets a new standard in the field, offering a robust solution for varied applications and paving the way for future advancements in face recognition technologies. In the future, we would enhance our adapter framework to address more complex image quality issues, such as varying lighting and obstructions. We aim to adapt our approach for real-time applications, broadening its utility in fields like surveillance and mobile authentication.

{
    \small
}


\newpage
\setcounter{page}{1}
\onecolumn
{
\centering
\Large
\textbf{\thetitle}\\
\vspace{0.5em}Supplementary Material \\
\vspace{1.0em}
\begin{figure}[htbp]
\centering
\includegraphics[width=\textwidth]{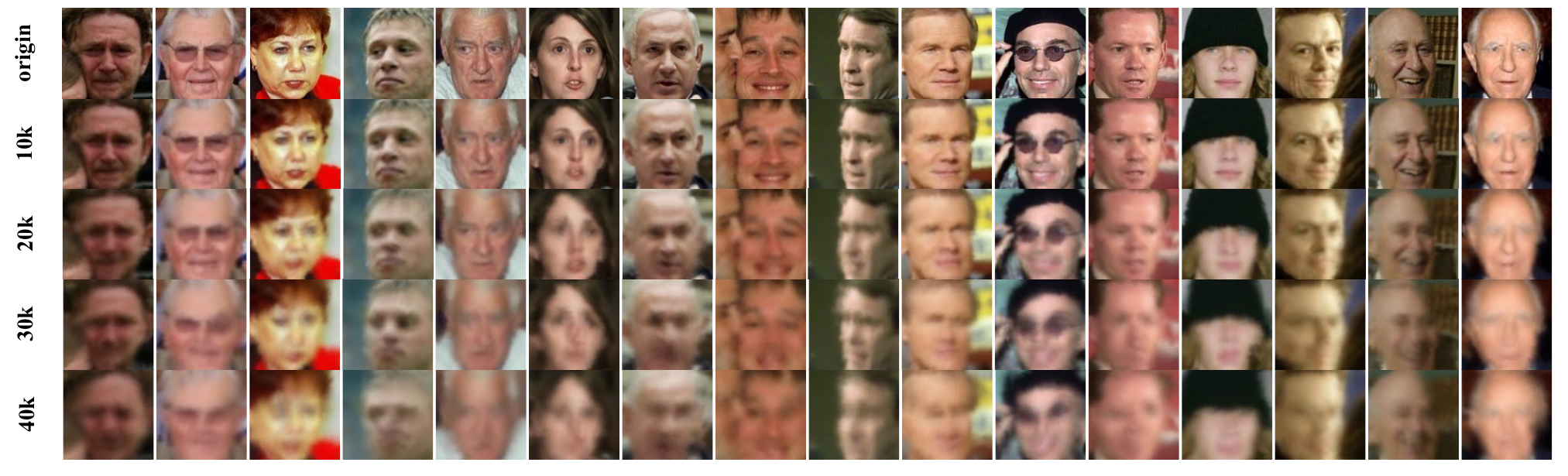}
\caption{Comparative analysis of image degradation under varying turbulence intensities. This figure presents a series of images processed using the TurbulenceSim tool, showcasing the effects of atmospheric turbulence at distinct parameters: $10k$, $20k$, $30k$, and $40k$ meters, as well as the origin images without degradation. Each row or set of images corresponds to one of these parameters, illustrating the progressive impact on image quality as the turbulence intensity increases.}
\label{fig:turbulence}
\end{figure}
}



\section{Image Processing in Synthetic Datasets}


\noindent \textbf{Image Degradation.} Image degradation refers to the process of quality reduction in an image, often encountered in real-world scenarios. Common causes of degradation include noise, blur, poor lighting, and compression artifacts. In the context of face recognition in the wild, these factors significantly challenge the recognition accuracy. Understanding the types and sources of degradation is crucial in developing robust face recognition systems. This includes, but is not limited to, Gaussian blur, which simulates out-of-focus images; salt-and-pepper noise, representing random pixel intensity disruptions; and motion blur, a result of rapid movement during image capture. Analyzing these degradation types helps in simulating real-world conditions for model training and evaluation.

In our study, we particularly focus on image degradation caused by atmospheric turbulence, which is a significant challenge in outdoor face recognition scenarios. Our methodology for simulating image degradation involves a sophisticated model of atmospheric turbulence, particularly focusing on the impact of this turbulence on long-distance face recognition. To simulate this type of degradation, we employ TurbulenceSim~\cite{chimitt_chan_sim}. TurbulenceSim realistically models the distortion effects caused by atmospheric turbulence, such as blurring and warping, which are common in images captured in long-range surveillance or outdoor environments. It replicates the atmospheric conditions by modeling the distortion effects like blurring and warping, which are typical in long-range surveillance or outdoor environments. The simulation involves multiple steps, as it is shown in Table~\ref{tab:turbulence_summary}.

Our image degradation model includes four distinct parameters: $meters = 10k, 20k, 30k, 40k$. These parameters represent the turbulence intensity in the algorithm, providing a range of scenarios from mild to severe atmospheric conditions. In Fig.~\ref{fig:turbulence}, we compare images processed under these different distinct parameters. This comparison demonstrates the varying effects of turbulence intensity on image quality. We use LFW~\cite{huang2008labeled}, CFP-FP~\cite{sengupta2016frontal}, and AgeDB~\cite{moschoglou2017agedb} as our evaluation dataset.



\begin{table*}[h]
\centering
\small
\resizebox{\linewidth}{!}{
\begin{tabular}{ccp{8cm}}
\hline 
\textbf{Serial} & \textbf{Name} & \textbf{Intro} \\
\hline
1 & Parameter Initialization & Involves setting up essential simulation parameters like image size, aperture diameter, propagation length, Fried parameter, and wavelength. \\
\hline 
2 & PSD Generation & Creates the PSD (Power Spectral Density) for tilt values, essential for simulating pixel shifts due to turbulence. \\
\hline 
3 & Image Tilt and Blur & Applies PSD to the input image to produce a tilted version, followed by the generation of a blurred image to mimic turbulence effects. \\
\hline 
4 & Zernike Coefficients & Generates Zernike coefficients for modeling wavefront distortions caused by atmospheric turbulence. \\
\hline 
5 & Processing Pipeline & Processes the images are through the turbulence model, first tilting and then blurring them to create a realistic dataset for face recognition challenges. \\
\hline
\end{tabular}
}

\caption{Summary of TurbulenceSim method steps}
\label{tab:turbulence_summary}
\end{table*}

\begin{figure*}[tp]
\centering
\includegraphics[width=\linewidth]{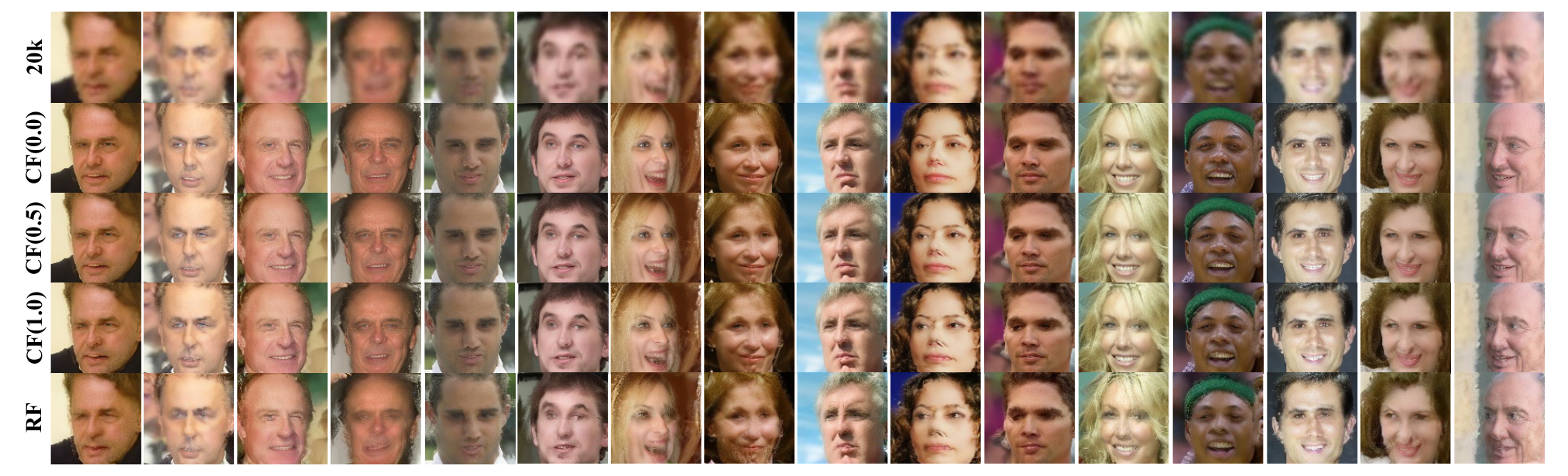}
\caption{Comparative evaluation of image restoration methods. This figure illustrates the effectiveness of different image restoration approaches on degraded facial images. CF refers to Codeformer, and RF refers to Restoreformer. For Codeformer, three distinct fidelity weight parameters ($w=0.0, 0.5, 1.0$) are evaluated, demonstrating how varying levels of fidelity weights impact the restoration quality. } 
\label{fig:restoration}
\end{figure*}

\noindent \textbf{Image Restoration.} In the process of restoring degraded images, our study employs two methodologies: Codeformer \cite{zhou2022codeformer} and Restoreformer \cite{wang2022restoreformer}. These tools enhance the quality of images affected by atmospheric turbulence, thereby generating the HQ images as one of the inputs of our method. In particular, there is a fidelity weight parameter w in Codeformer, which we vary across three different settings: \( w=0.0, 0.5, \) and \( 1.0 \). The fidelity weight \( w \) lies in the range \([0, 1]\), where a smaller \( w \) value generally produces a higher-quality result, and a larger \( w \) value yields a result with higher fidelity to the original degraded image. This flexibility allows us to balance between enhancing image quality and maintaining the authenticity of the original features.

In our experimental setup, we compare the performance of images restored using Codeformer with the three different fidelity weight parameters and Restoreformer. This comparison is important to understand how different restoration methods and parameter settings affect the overall quality and fidelity of the restored images. It also provides insights into the optimal restoration strategy for enhancing the performance of face recognition systems in wild scenarios. Fig. \ref{fig:restoration} presents a comprehensive comparison of images restored using these methods. It showcases the distinct outcomes of each setting of Codeformer and the results achieved using Restoreformer.


\section{Fusion Structure}

\begin{table}
\centering

\begin{tabular}{ccccc}
\hline Param  & d\_model & nhead & dropout & bias \\
\hline Value & $512$ & $8$ & $0.0$ & True \\

\hline
\end{tabular}

\caption{This table enumerates the specific parameters used in the nn.MultiheadAttention layer within both Cross-Attention and Self-Attention components of our Fusion Structure. The parameters include d\_model (dimension of the model), nhead (number of attention heads), dropout rate, and whether bias is included.}
\label{tab:multihead}
\end{table}

\begin{figure*}[h]
\centering
\includegraphics[width=0.9\linewidth]{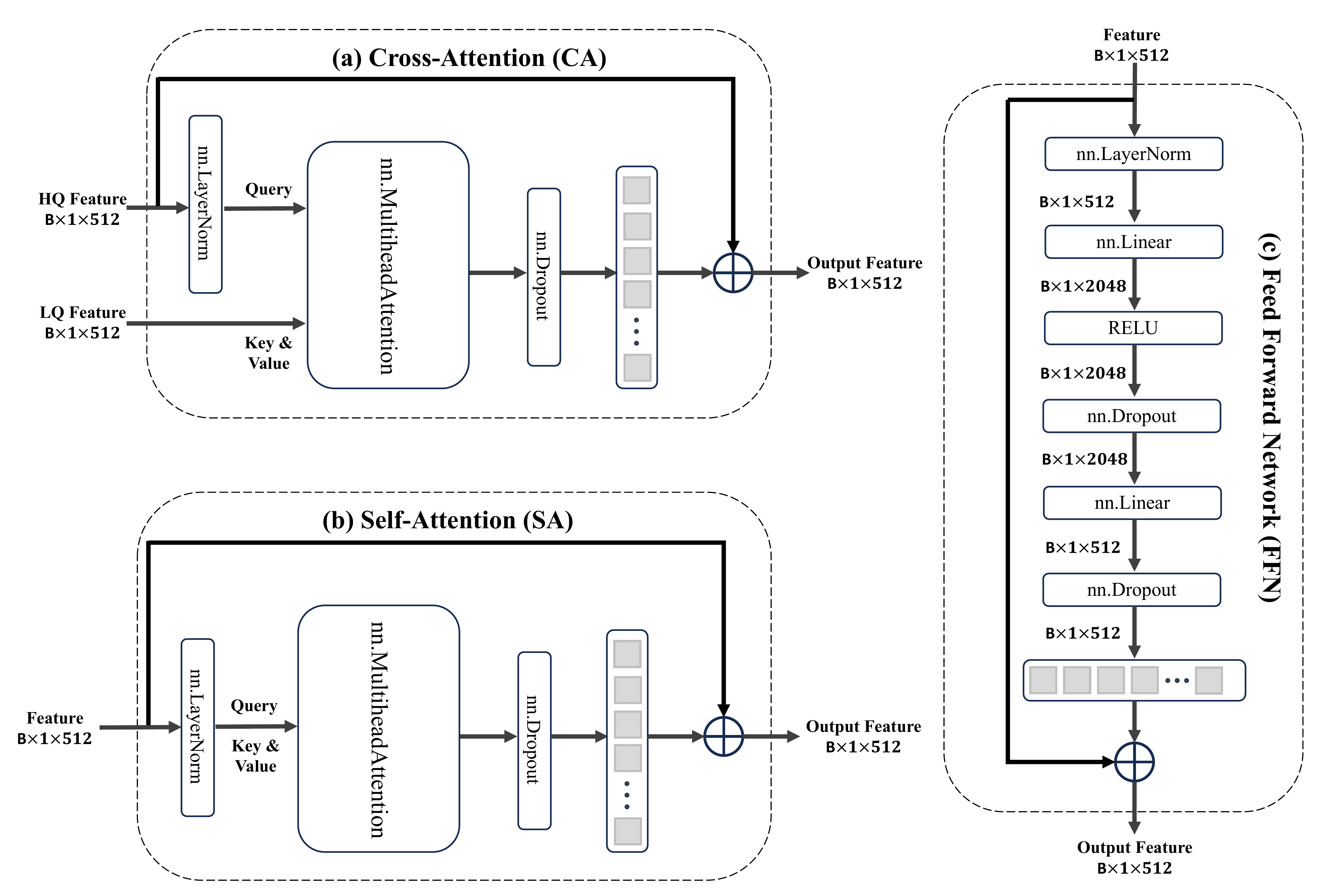}
\caption{\textbf{Detailed Illustration of Fusion Structure Components.} This figure provides a visual representation of the three key parts of our Fusion Structure: (a) Cross-Attention, (b) Self-Attention, and (c) Feed Forward Network. Each diagram in the figure meticulously details the layers involved, along with the inputs and outputs of these respective components. In both (a) and (b), we use the same parameters to initialize the method nn.MultiheadAttention, as it is shown in Table \ref{tab:multihead}. The Cross-Attention diagram (a) showcases how features from different image streams are integrated, while the Self-Attention diagram (b) illustrates the processing of features within a single image stream. Lastly, the Feed Forward Network diagram (c) depicts the sequential layers responsible for refining and preparing the final feature output.} 
\label{fig:fusion}
\end{figure*}

The Fusion Structure in our study plays an important role in integrating and processing features from both restored and original degraded images. The \href{https://github.com/facebookresearch/Mask2Former}{codebase} for this model is primarily derived from \cite{cheng2021maskformer, cheng2021mask2former}.

\noindent \textbf{Cross-Attention.} As it is shown in Fig. \ref{fig:fusion} (a), Cross-Attention is designed to integrate features from both the restored and original image streams. This integration is essential for the model to effectively understand and interpret faces in varying conditions. By aligning and correlating features from the dual inputs, Cross-Attention ensures that the combined feature set is and comprehensive, enhancing the face recognition system's ability to adapt to diverse image qualities.

\noindent \textbf{Self-Attention.} As it is shown in Fig. \ref{fig:fusion} (b), Self-Attention is utilized to process the input features. By modeling and associating these features, the mechanism captures the global contextual information of the image. This approach enables the model to consider the entire image's contextual information. Consequently, richer features essential for effective face recognition are extracted, thereby improving the accuracy of the system.

\noindent \textbf{Feed Forward Network.} As it is shown in Fig. \ref{fig:fusion} (c), Feed Forward Network (FFN) acts as the final processing stage in the network, refining the features post-attention processing. It consists of multiple dense layers that further enhance the feature set, ensuring that the final output is well-suited for the face recognition task.

\section{Results of Visualization}

\begin{figure*}[h]
\centering
\includegraphics[width=\linewidth]{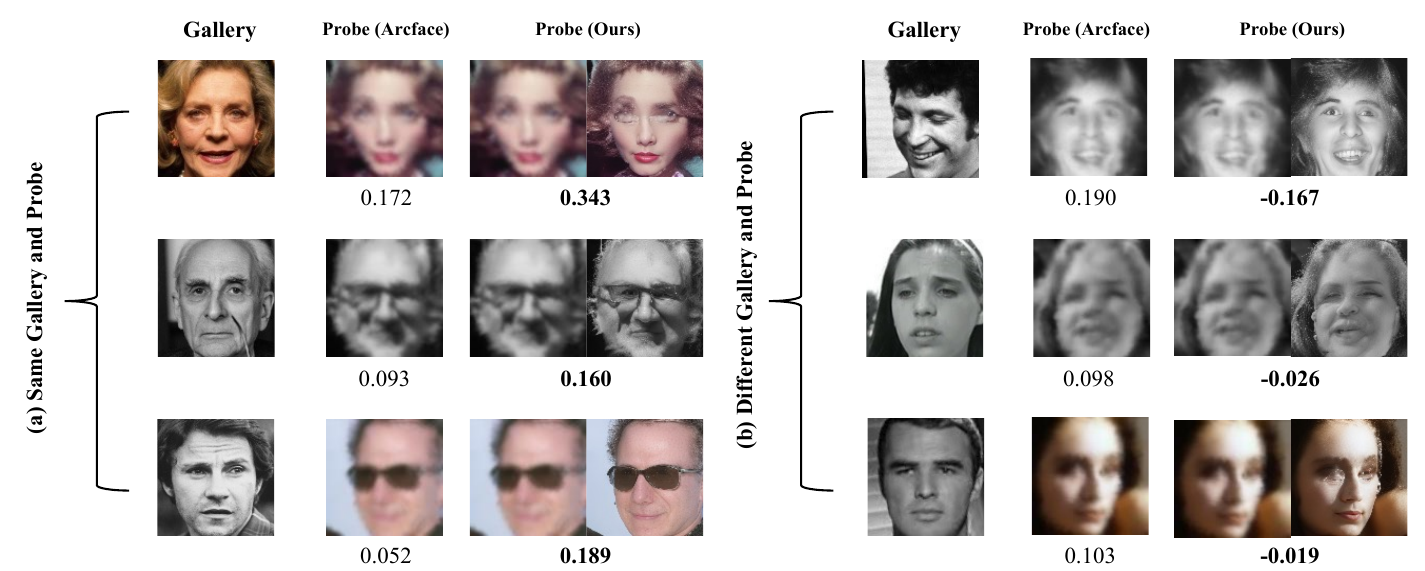}
\caption{Comparative analysis of face recognition methods on image pair. (a) illustrates the comparison results using different methods on an image pair of the same person, while (b) presents the results for an image pair of different people. Each image pair is accompanied by a cosine similarity score, ranging from $-1$ to $1$, where a higher score indicates greater similarity between the images, and vice versa.} 
\label{fig:restoration-A}
\end{figure*}

\begin{figure*}[h]
\centering
\includegraphics[width=\linewidth]{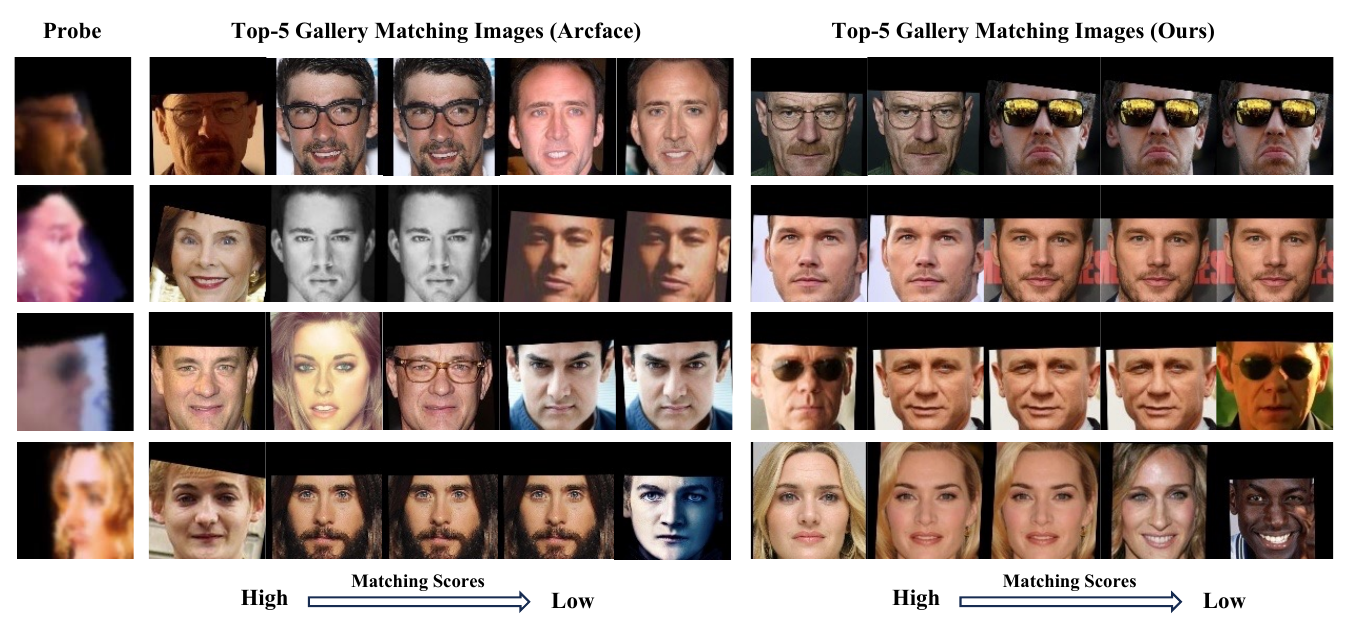}
\caption{Comparative analysis of face recognition methods on $1:N$ matching. This figure displays the Top 5 matching results for blurred face probes using both Arcface and our method. Each line contains an input probe and the Top 5 matches of the gallery using each of the two methods. } 
\label{fig:restoration-B}
\end{figure*}

In this section, we present the results of our visualization experiments, designed to compare the performance of the traditional Arcface \cite{deng2019arcface} method with our method in face recognition, especially when dealing with low-quality images. These visualization experiments are crucial for demonstrating the practical effectiveness of our method in real-world scenarios where image quality often varies.

For the probe images in our experiments, we employ the parameter TurbulenceSim \( m = 20k \) to blur the images. Subsequently, we use Codeformer with a fidelity weight \( w = 0.5 \) to restore the images.

Our comparison is based on the use of cosine similarity as a metric to determine the degree of similarity between the gallery feature and the probe feature. The formula for calculating the cosine similarity of two vectors \( \mathbf{s_1} \) and \( \mathbf{s_2} \) is defined as:

\begin{equation}
\text{Cosine Similarity} (\mathbf{s_1}, \mathbf{s_2}) = \frac{\mathbf{s_1} \cdot \mathbf{s_2}}{\|\mathbf{s_1}\| \|\mathbf{s_2}\|}
\end{equation}

Where \( \cdot \) denotes the dot product of the two vectors, and \( \|\mathbf{s_1}\| \) and \( \|\mathbf{s_2}\| \) are the magnitudes of vectors \( \mathbf{s_1} \) and \( \mathbf{s_2} \), respectively.

\noindent \textbf{Image Pair Score Comparison.} In this experiment, we delve into a detailed comparative analysis between the Arcface method and our proposed method using a $1:1$ image pair approach. This comparison is visually represented in Fig. \ref{fig:restoration-A}. In this experimental setup, pairs of images are used to evaluate the effectiveness of both methods in recognizing and verifying the same individual. Each pair consists of two images, one serving as a gallery and the other as a probe.

Our experimental results demonstrate that the features extracted by our method are more representative of the key characteristics of the face. This enhanced representation leads to a more accurate determination of whether the two images in a pair represent the same person.

\noindent \textbf{Top-K People.} In this experiment, we explore the comparison between the Arcface method and our proposed approach using a $1:N$ methodology. This approach is key to understanding the performance of face recognition methods in scenarios where a single input (probe) face is compared against a larger database (gallery) of faces. 

In our experimental setup, we introduce blurred input face probes. These probes are then matched against a pre-established gallery of original, high-quality face images. Fig. \ref{fig:restoration-B} presents the results of these experiments, showing the Top 5 matches for each probe image as identified by both methods. The experimental data show that our method consistently ranks the correct individual higher in the list of potential matches compared to Arcface, thereby underlining its efficacy in practical, real-world face recognition applications where image quality can vary significantly.



%
%

\bibliographystyle{splncs04}
\bibliography{main}
\end{document}